\documentclass[a4paper,11pt,fleqn]{article}
\usepackage{pstricks}
\usepackage{amsmath}
\usepackage{amssymb}
\usepackage{mathrsfs} %pour mathscr
\usepackage{booktabs}
\usepackage{euscript}
\usepackage{exscale,relsize}
\usepackage{graphicx}
\usepackage{booktabs}
\usepackage{srcltx}
\usepackage{xcolor}
\usepackage{charter}
\usepackage{url}
\newcommand{\email}[1]{\href{mailto:#1}{\nolinkurl{#1}}}
\PassOptionsToPackage{normalem}{ulem}
\usepackage{ulem}

\renewcommand{\leq}{\ensuremath{\leqslant}}

\newcommand{\argmin}{\ensuremath{\text{\rm argmin}\,}}

\numberwithin{equation}{section}
\definecolor{labelkey}{rgb}{0,0.08,0.45}
\definecolor{refkey}{rgb}{0,0.6,0.0}
\definecolor{Brown}{rgb}{0.45,0.0,0.05}
\definecolor{dgreen}{rgb}{0.00,0.49,0.00}
\definecolor{dblue}{rgb}{0,0.08,0.75}
\RequirePackage[colorlinks,hyperindex]{hyperref}
\hypersetup{linktocpage=true,citecolor=dblue,linkcolor=dgreen}
\title{\sffamily\LARGE A Bridge Between Hyperparameter Optimization \\ and Learning-to-learn}
\date{~}

\author{
  Luca Franceschi$^{1, 2, *}$, Paolo Frasconi$^{3}$, Michele Donini$^{1}$, Massimiliano Pontil$^{1, 2}$  \\ \\
  $^{1}$Istituto Italiano di Tecnologia  \vspace{.1truecm} \\
  $^{2}$Department of Computer Science, University College London\vspace{.1truecm}   \\
  $^{3}$Universit\`a  degli Studi di Firenze \\ \\
  $^{*}$\texttt{luca.franceschi@iit.it}
}

\usepackage{floatrow}
\newfloatcommand{capbtabbox}{table}[][\FBwidth]

\usepackage{amsmath,amsthm,amssymb}
\usepackage{algorithm,graphicx}
\usepackage{hyperref,mathrsfs}
\usepackage{xcolor}

\usepackage[fleqn,tbtags]{mathtools}
\usepackage{natbib}

\newlength\Colsep
\allowdisplaybreaks

\usepackage{varwidth}
\def\RSet{\mathbb{R}}
\DeclareMathOperator{\E}{\mathbb{E}}
\usepackage[status=draft,margin=false,inline=true,index=true]{fixme}

\usepackage{calrsfs}
\DeclareMathAlphabet{\omathcal}{OMS}{zplm}{m}{n}

\begin{document}
% \nipsfinalcopy is no longer used

\maketitle

\begin{abstract}
We consider a class of a nested optimization problems involving inner and outer objectives. We observe that by taking into explicit account the optimization dynamics for the inner objective it is possible to derive a general framework that unifies gradient-based hyperparameter optimization and meta-learning (or learning-to-learn). Depending on the specific setting, the variables of the outer objective take either the meaning of hyperparameters in a supervised learning problem or parameters of a meta-learner. We show that some recently proposed methods in the latter setting can be instantiated in our framework and tackled with the same gradient-based algorithms. Finally, we discuss possible design patterns for learning-to-learn and present encouraging preliminary experiments for few-shot learning.

\end{abstract}

\section{Introduction and framework}  \label{sec:framework}

Hyperparameter optimization \citep[see, e.g.,][]{moore_model_2011,bergstra_algorithms_2011,bergstra_random_2012,maclaurin_gradient-based_2015,bergstra_making_2013,hutter_beyond_2015,franceschi_forward_2017} 
 is the problem of tuning the value of certain parameters 
% (such as regularization coefficients, learning rate, etc.) 
that control the
behaviour of a learning algorithm. This is typically
obtained by minimizing the expected error 
% with respect to the
w.r.t. the hyperparameters, using the empirical loss on a validation set as a
proxy. Meta-learning
\citep[see, e.g.,][]{thrun_learning_2012,baxter_theoretical_1998,Maurer,MPR,vinyals_matching_2016,santoro_meta-learning_2016,ravi_optimization_2016,mishra_meta-learning_2017,finn_model-agnostic_2017} 
is the problem of 
%automatically synthesizing 
inferring a 
%supervised 
learning
algorithm from a collection of datasets 
% changed LUCA
%in order to obtain boosted performances when applied to novel 
%(but related)
%try again
in order to obtain good performances on unseen
datasets.
% by leveraging knowledge acquired from previously 
% %seen problems.
% solved problems. % instances.
% end 
% so that fitting the generated algorithm
% on novel datasets it will leverage
% knowledge acquired from the previously seen datasets, resulting in
% boosted prediction performance. 
% This is particularly useful when the
% novel datasets are very small (i.e., in the few-shots learning
% setting). 
Although hyperparameter optimization and meta-learning are different and apparently unrelated
problems,
% and although the relevant bodies of literature do not
% largely overlap, 
they can be both formulated as special cases of a
wider framework that we will introduce. This connection and our observations on learning-to-learn represent the main contribution of this work.
%
% that involves a bilevel optimization problem at its
% core, and whose inner objective results from the iteration of a
% parametric dynamical system.

We start by considering bilevel optimization problems \citep[see e.g.][]{colson2007overview}
of the form
\begin{equation}
\min_{\lambda\in \Lambda}  f(\lambda)
\label{eq:1}
\end{equation}
where $\Lambda\subseteq \RSet^m$ and
\begin{equation}
f(\lambda) = \inf_w \{E(w, \lambda ) : w \in {\rm argmin}_u L_{\lambda}(u)\}.
\label{eq:2}
\end{equation}
%\footnote{Meglio lascare che $E$ dipenda esplicitamente da $\lambda$}
% As we shall see, specific instances of this problem include hyperparamter optimization and meta-learning. 
We will call the function $f:\Lambda\to\RSet$ the \emph{outer
objective} (or outer loss),
and, for every $\lambda \in \Lambda$,  $L_{\lambda}:\RSet^d\to\RSet$ is called the \emph{inner objective} (or inner loss). Note that $\{L_{\lambda} \;:\; \lambda \in \Lambda\}$ is a class of objectives parametrized by $\lambda$.
As prototypical example of \eqref{eq:2} consider the case that $L_{\lambda}$ is
a regularized empirical error for supervised learning, $E$ is an (unregularized) validation error, $\lambda$ a regularization parameter and $w$ the
parameters of the model.

% In general problem \eqref{eq:1} may not have a solution \citep{colson2007overview} but we assume this is in not an issue here. 
% LUCA IS THIS SENTENCE TRUE? WHAT ARE THE NECESSARY CONDITIONS FOR EXISTENCE?
Following \citep{domke_generic_2012,maclaurin_gradient-based_2015,franceschi_forward_2017} % IMPORT piu' citazioni Pedregosa.
we
%==\luca{Teniamo approximate???}
approximate the solutions of problem \eqref{eq:1} by replacing the ``argmin" in problem \eqref{eq:2} by the $T$-th iterate of a dynamical system 
%obtained by an iterative system 
of the form
\begin{equation}
  \label{eq:dynamics}
  w_0 = \Phi_0(\lambda); \quad w_t = \Phi_t(w_{t-1},\lambda) \ \ t=1,\dots,T
\end{equation}
where $T$ is the number of iterations, $\Phi_0:\RSet^m\to\RSet^d$ is a smooth initialization mapping and,
for every $t \in \{1,\dots,T\}$,
$
\Phi_t : \RSet^d \times \RSet^m \rightarrow \RSet^d
$
is a smooth mapping that represents the operation performed by the
$t$-th step of an optimization algorithm. Since the algorithm might involve auxiliary variables $v$, e.g. velocities when using stochastic gradient descent with momentum (SGDM), we replace $w$ with a state vector $s=(w,v)$.
Using this notation, we 
% approximate problem \eqref{eq:1} by the constrained optimization problem
formulate the following constrained optimization problem
\begin{equation}
  \begin{array}{cl}
    \min\limits_{\lambda,s_1,\dots,s_T} & f(\lambda) = E(s_T, \lambda) \\ 
   \text{~~~~~subject to} & 
    s_0 = \Phi_0(\lambda) \\
   & s_t =  \Phi_t(s_{t-1},\lambda),~t \in \{1,\dots,T\}.
  \end{array}
  \label{eq:general:constrained}
\end{equation}
This reformulation of the original problem allows for an efficient computation of the gradient of $f$, either in time or in memory \citep{maclaurin_gradient-based_2015, franceschi_forward_2017}, by making use of Reverse or Forward mode algorithmic differentiation \citep{griewank2008evaluating}. 
%[CITE]\footnote{Parlare qui brevemente degli algoritmi e delle relative versioni online?}. 
Moreover, by considering explicitly the learning dynamics, it is possible to compute the hypergradient with respect to the hyperparameters that appear inside the optimization dynamics (e.g. step size or momentum factor if $\Phi$ is SGDM), as opposed to other methods that compute the hypergradient at the minimizer of the inner objective \citep{pedregosa2016hyperparameter}. This key fact allows for the inclusion of learning-to-learn, more specifically learning-to-optimize, into the framework. 
%\textbf{Paolo: non mi torna tanto, LtO non Ã¨ un caso speciale di LtL secondo quanto ci siamo detti varie volte, forse toglierei more specifically..}
% 
% LUCA: CHIEDERE A MASSI 
%A related bilevel programming formulation of HO was originally proposed in~\citep{moore_model_2011} for the case of linear SVM (whose inner objective is convex).
In the next two sections we show that gradient-based hyperparameter optimization and learning-to-learn share this same latter underlying mathematical formulation.

\iffalse We highlight the generality of the framework. The vector of
hyperparameters $\lambda$ may include components associated with the
training objective, and components associated with the iterative
algorithm. For example, the training objective may depend on
hyperparameters used to design the loss function as well as multiple
regularization parameters. Yet other components of $\lambda$ may be
associated with the space of functions used to fit the training
objective (e.g. number of layers and weights of a neural network,
parameters associated with the kernel function used within a kernel
based method, etc.). The validation error $E$ can in turn be of
different kinds. 
We may however consider multiple validation objectives, in that the
hyperparameters associated with the iterative algorithm ($\mu$ and
$\gamma$ in the case of momentum mentioned above) may be optimized
using the training set, whereas the regularization parameters would
typically require a validation set, which is distinct from the
training set (in order to avoid over-fitting).  
\fi

\section{Gradient-based hyperparameter optimization} \label{sec:HO}

In the context of hyperparameter optimization, we are interested in minimizing the generalization error of a model $g:\omathcal{X}\to\omathcal{Y}$, parametrized by a vector $w$, with respect to $\lambda$.
%\footnote{as opposed to learning-to-learn scenario, where the "highlight" is on the meta-learner, and thus on the "hyperprameters" by themselves} 
The outer optimization variables are in this context called hyperparameters and the outer objective is generally an empirical validation loss. 
Specifically, a set of labelled examples $D=\{z_i\}_{i=1}^n$, where $z_i = (x_i,y_i) \in\omathcal{X}\times\omathcal{Y}$, is spit into training and validation sets $D_{\operatorname{tr}}$, $D_{\operatorname{val}}$. 
%(a third spit $D_{\operatorname{test}}$ is held out to compute the final performance of the learned model $g$). 
The inner objective is computed on (mini-batches of) examples from $D_{\operatorname{tr}}$ while the outer objective, that represents a proxy for the generalization error of $g$, is computed on $D_{\operatorname{val}}$.
Assuming, for simplicity, that the optimization dynamics is given by stochastic gradient descent, and thus that the state $s=w$, problem \eqref{eq:general:constrained} becomes
\begin{equation}
  \begin{array}{cl}
    \min\limits_{\lambda,w_1,\dots,w_T} & f(\lambda) = 
    %\frac{1}{|D_{\operatorname{val}}|} 
    \sum\limits_{z\in D_{\operatorname{val}}} E(w_T, z) \\ 
   \text{~~~~~subject to} & 
    w_0 = \Phi_0(\lambda) \\
   & w_t =  w_{t-1} - \eta
   %\frac{\eta}{|B_t|} 
  \sum\limits_{z\in B_t} \nabla L_{\lambda}(w_{t-1}, z), \quad~t \in \{1,\dots,T\},
  \end{array}
  \label{eq:general:ho}
\end{equation}
where $B_t\subset D_{\operatorname{tr}}$ is a mini-batch of samples at the $t$-th iteration, $\eta$ is a learning rate (a component of $\lambda$) and where we made explicit the dependence of the loss functions on the examples. In this setting, the outer loss $E$ does not depend explicitly on the hyperparameters $\lambda$.
The above formulation allows for the computation of the hypergradient of any real valued hyperparameter, so that hyperparameters can be optimized with a gradient descent procedure. 
%While the requirement that $\lambda\in\RSet^m$ can be rather strict, since in practice many hyperparameters take values on a discrete sets (e.g. number of layers of a neural network, whether to perform or not a certain data preprocessing, ...), 
Having access to hypergradients makes it feasible to optimize a number of hyperparameters of the same order of that of parameters, a situation which arise in the setting of learning-to-learn. 
%Moreover, gradient informations can be used to boost the effectiveness of Bayesian optimization methods \citep{wu2017bayesian}.

Since in this context the total number of iterations might be often high due to large datasets or complex models, to speed up the  optimization and to reduce memory requirements, it is possible to compute partial hypergradients at intermediate iterations, either in reverse or forward mode, and update $\lambda$ online several times before reaching the final iteration $T$ \citep{franceschi_forward_2017}.  

%As a simple example, the dynamical system \eqref{eq:dynamics} may be gradient descent to minimize a regularized empirical error $L$ in supervised learning, and the parameters $\lambda \in \RSet^2$ may be formed by the regularization parameter and the stepsize.

\section{Learning-to-learn}  \label{sec:l2l}

The aim of meta-learning is 
%a lifting of a standard machine learning problem: the task is 
to learn an algorithm capable of solving ground learning problems originated by a (unknown)  
distribution $\mathcal{P}$. A meta-dataset $\omathcal{D}=\{D^j\}_{j=1}^{N}$ is thus a 
%(possibly infinite) 
collection of datasets, or \emph{episodes}, sampled from $\mathcal{P}$, where each dataset $D^j=\{z^j_i\}_{i=1}^{n_j}$ with $z_i^j = (x_i^j,y_i^j)\in\omathcal{X}^j\times\omathcal{Y}^j$ is linked to a specific task. 
%\luca{parlare anche di meta-validation e meta-test sets?} 
We are interested in learning an algorithm capable of ``producing'' ground models $g^j:\omathcal{X}^j\to\omathcal{Y}^j$, which we assume identified by parameter vectors $w^j$. 
The algorithm itself can be thought of as a meta-model $q$, or \emph{meta-learner}, 
parametrized by a vector $\lambda$, so that $w^j = q(D^j, \lambda)$. The meta-learner $q:\omathcal{D}\to\omathcal{W}$ is viewed as a function which maps datasets to models (or weights), effectively making it a (non-standard, usually highly parametrized) learning algorithm. As a learning dynamics, in general, the meta-model can act in an iterative way, so that $q = q_T \circ q_{T-1} \circ \dots \circ q_0$. 
Moreover, like the case of a standard optimization algorithm, the meta-learner can make use of auxiliary variables $v^j$, forming state vectors $s^j=(w^j, v^j)$. Since the ground models should exhibit good generalization performances on their specific task, each dataset $D^j$ can be split into training and validation\footnote{Note that some authors \citep[e.g.][]{ravi_optimization_2016} refer to this latter set as the test set.} 
sets $D_{\rm{tr}}^j, D_{\rm{val}}^j$, and 
$q$ can be trained to minimize the average validation error over tasks, which constitutes a natural outer objective in this setting. For each task, the meta-learner produces a sequence of states $s^j_0 = q_0(D^j_{\operatorname{tr}}, \lambda), \dots, s^j_T = q_T(D^j_{\operatorname{tr}}, s^j_{T-1}, \lambda)=q(D^j_{\operatorname{tr}}, \lambda)$.
%MMM so q is the learning algorithm!

We can thus formulate problem \eqref{eq:general:constrained} for learning-to-learn as follows:
\vspace{-.2cm}
\begin{equation}
  \begin{array}{cl}
    \min\limits_{\lambda,s_0^1,\dots,s_T^N} & f(\lambda) = 
    \sum\limits_{j=1}^{N}
    \frac{1}{|D^j_{\operatorname{val}}|} 
    \sum\limits_{z\in D^j_{\operatorname{val}}} 
    E^j(s^j_T, \lambda, z) \\ 
   \text{~~~~~subject to} & s^j_0 = q_0(D^j_{\operatorname{tr}}, \lambda) \\
   & s_t^j =  q_t(D^j_{\operatorname{tr}}, s_{t-1}^j, \lambda) \quad~j \in \{1,\dots,N\}, \; t\in\{1,\dots,T\},
  \end{array}
  \label{eq:general:l2l:form1}
\end{equation}
\vspace{-.05cm}
where the functions $E^j$ are task specific losses. The meta-model plays the role of the mapping $\Phi$ in \eqref{eq:dynamics}, thus reducing the problem of learning-to-learn to that of \emph{learning a training dynamics}, or its associated parameters $\lambda$. The meta-learner parameters mirror the hyperparameters in the context of hyperparameter optimization in Section \ref{sec:HO} and can be optimized with a gradient descent procedure on the outer objective. The inner objective does not appear explicitly in problem \eqref{eq:general:l2l:form1}, but we assume that the meta-learner has access to task specific inner objectives $L^j$.
% COMMENT: or not task specific. L can depend also on states of different ground models... like in the case of MTL (?). this may constitute a (kind of minor) difference between the two setting, in the sense that usually in l2l task interaction is only trough meta model (and hyperparameters) while in MTL is contained in the loss...

While in principle $q$ could be implemented by any parametrized mapping, the design of meta-learning models can follow three non-exclusive natural directions:

$\bullet$ \emph{Learning-to-optimize}: $q$ can replace a gradient-based optimization algorithm \citep{andrychowicz_learning_2016, wichrowska_learned_2017}, acting on the weights of ground models as $w^j_{t+1} = w^j_t - {q_t}(B^j_t, s^j_{t-1}, L^j, \nabla_{w} L^j)$, where $B^j_t\subseteq D_{\operatorname{tr}}^j$ is a mini-batch of examples.
The meta-model is often interpreted \citep{ravi_optimization_2016} as a recurrent neural network, whose hidden states $v^j$ are the analogue of auxiliary variables in Section \ref{sec:framework}. 
Alongside the update rule, it is possible to learn an initialization for the ground models weights, described by the mapping $q_0$. For instance, \citep{finn_model-agnostic_2017} set $q_0(D^j_{\operatorname{tr}}, \lambda)=\lambda=w^j_0$ assuming that all the input and output spaces of the tasks in $\omathcal{D}$ have the same dimensionality, and use gradient descent for the following steps;
%\begin{equation}
%\left( \begin{array}{c}
%  \theta_{t+1} \\ h_{t+1} \end{array} \right)  = 
%  \left( \begin{array}{c}
%  \theta_{t} + m_1(\theta_t, h_t, \phi) \\ 
%   m_2(\theta_t, h_t, \phi)
%  \end{array}
%  \right)  
%\end{equation}

\noindent
$\bullet$ \emph{Learning meta-representations}: the meta-learner is composed by a gradient descent procedure and a mapping from ground task instances $x$ to intermediate representations $h(x, \lambda)\in\omathcal{Z}$. In this case the ground models are mappings $g^j:\omathcal{Z}\to\omathcal{Y}^j$ and an update on ground model weights is of the form  $w^j_{t+1} = w^j_t - \eta \sum_{(x,y)\in D_{\operatorname{tr}}^j}\nabla L^j(w^j_{t-1}, h(x, \lambda), y)$. This approach can prove particularly useful in cases where the instance spaces are structurally different among tasks. It differs from standard representation learning in deep learning \citep{bengio2009learning,Goodfellow-et-al-2016} since the meta-training loss is specifically designed for promoting generalization across tasks;

\noindent
$\bullet$ \emph{Learning ground loss functions}: the meta-learner can be a gradient descent algorithm that optimize a learned inner objective. 
%MASSI: ho commentato questo - mi e' parso poco chiaro 
%That is, we consider the dynamics $w^j_{t+1} = w^j_t - \eta \nabla \ell(B^j_t, w^j_{t-1}, \lambda)$, where $\ell$ is a parametrized class of loss functions (including a possible regularizer). 
%As an example consider multi-task learning where $w=(w^j)_{j=1}^{N}$ is the joint parameter vector and $D_{\operatorname{tr}}=\cup D^j_{\operatorname{tr}}$ is the complete training set, then $h(D_{\operatorname{tr}}, w, \lambda) = L(D_{\operatorname{tr}}, w) + R(w, \lambda)$, where $R$ is a multi-task regularizer such as $\sum_{i\neq j} \lambda_{ij}||w_i - w_i||^2$ in the case of linear models. Yet, 
For example, we may directly parametrize the training error $L$ (which in a standard supervised learning setting is usually a mean squared error for regression or a cross-entropy loss for classification), or learn a multitask regularizer which provides a coupling among different learning tasks 
%a meta-error function adapted to the tasks 
%
in $\mathcal{P}$. 
%\luca{Qui magari sarebbe bello anche scrivere una frase di collegamento con GAN in cui il discriminator funge da training error per il generator.}    
%MASSI: terrei l'esempio anche se e'un po' vago/qualitativo. Forse il link a learning the muit-task regularizer lo rende piu' concreto? ho provato a menzionare questo brevemente
%\textbf{Paolo: Non so se metterei il terzo caso dato che non lo fa nessuno e non abbiamo esperimenti --- peraltro i design patterns si scoprono, non si inventano...}

In the next section we presents experiments that explore the second design pattern. For experiments on gradient-based hyperparameter optimization we refer to \citep{franceschi_forward_2017}.
%  where we optimize the example weights of a weighted loss for noisy labels detection, learned the coefficient of a multi-task regularizer and optimized algorithmic and design hyperparameters of a large-scale phone recognition model with an online algorithm. 

\section{Experiments} \label{sec:exp}

We report preliminary results on the problem of few-shots learning,  using MiniImagenet \citep{vinyals_matching_2016},
%, a recently affirmed benchmark dataset for few-shot learning. 
a subset of ImageNet \citep{deng2009imagenet}, that contains 60000 downsampled images from 100 different classes. As in  \citep{ravi_optimization_2016}, we build meta-datasets by sampling ground classification problems with 5 classes, where each episode $D=(D_{\operatorname{tr}},D_{\operatorname{val}})$ is constructed so that $D_{\operatorname{tr}}$ contains 1 (one-shot learning) or 5 (5-shots learning) examples per class and $D_{\operatorname{val}}$ contains 15 examples per class. Out of 100 classes, 64 classes are included in a training meta-dataset $\omathcal{D}_{\operatorname{tr}}$ from which we sample datasets for solving problem \eqref{eq:general:l2l:form1}; 16 classes form a validation meta-dataset $\omathcal{D}_{\operatorname{val}}$ which is used to tune meta-learning hyperparameters while a third meta-dataset $\omathcal{D}_{\operatorname{test}}$ with the remaining 20 classes is held out for testing. 
We use the same split and images proposed by  \citep{ravi_optimization_2016}. The code is available at  \url{https://github.com/lucfra/FAR-HO}.

\begin{figure}[h]
\CenterFloatBoxes
\begin{floatrow}
\capbtabbox{
 \begin{tabular}{lcc}
      \hline
  %     \abovespace\belowspace
      5-classes accuracy \% & 1-shot & 5-shots  \\ \hline \hline 
      \emph{Fine-tuning}            & $28.86 \pm 0.54$  &$49.76 \pm 0.79$     \\
      \emph{Nearest-neighbor}           & $41.08 \pm 0.70$ & $51.04 \pm 0.65$      \\
      \emph{Matching nets} & $43.44 \pm 0.77$ & $55.31 \pm 0.73$  \\
      \emph{Meta-learner LSTM} & $43.56 \pm 0.84$ & $60.60 \pm 0.71$ \\
      \emph{MAML} & $48.70\pm 1.75$ & $63.11 \pm 0.92 $ \\
      \emph{Hyper-Repr.} (ours) & $47.01 \pm 1.35$ & $61.97 \pm 0.76 $   \\  \hline 
    \end{tabular}
}{%
  \caption{\small{Mean accuracy scores with 95\% confidence intervals, computed on episodes from %$D^j_{\operatorname{val}}\in
  $\omathcal{D}_{\operatorname{test}}$, of various methods on 1-shot and 5-shot classification problems on MiniImagenet. \label{tab:results}}}%
}
% \hspace{-1cm}
\ffigbox{
\hspace*{-1.5cm}
  \includegraphics[width=.35\textwidth]{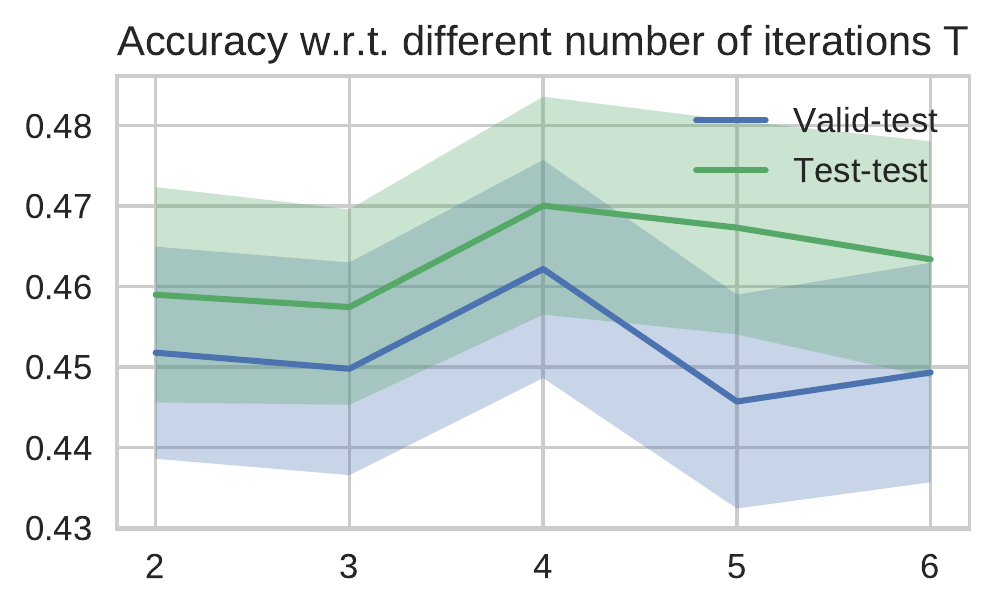}
}{% 
\vspace*{-.2cm}
  \caption{\small{Meta-validation of the number \newline of gradient descent steps on ground models \newline parameters $T$ for one-shot learning.\label{fig:updates}}}%  
}
\end{floatrow}
\end{figure}

\begin{figure}[h] 
\includegraphics[width=0.48\textwidth]{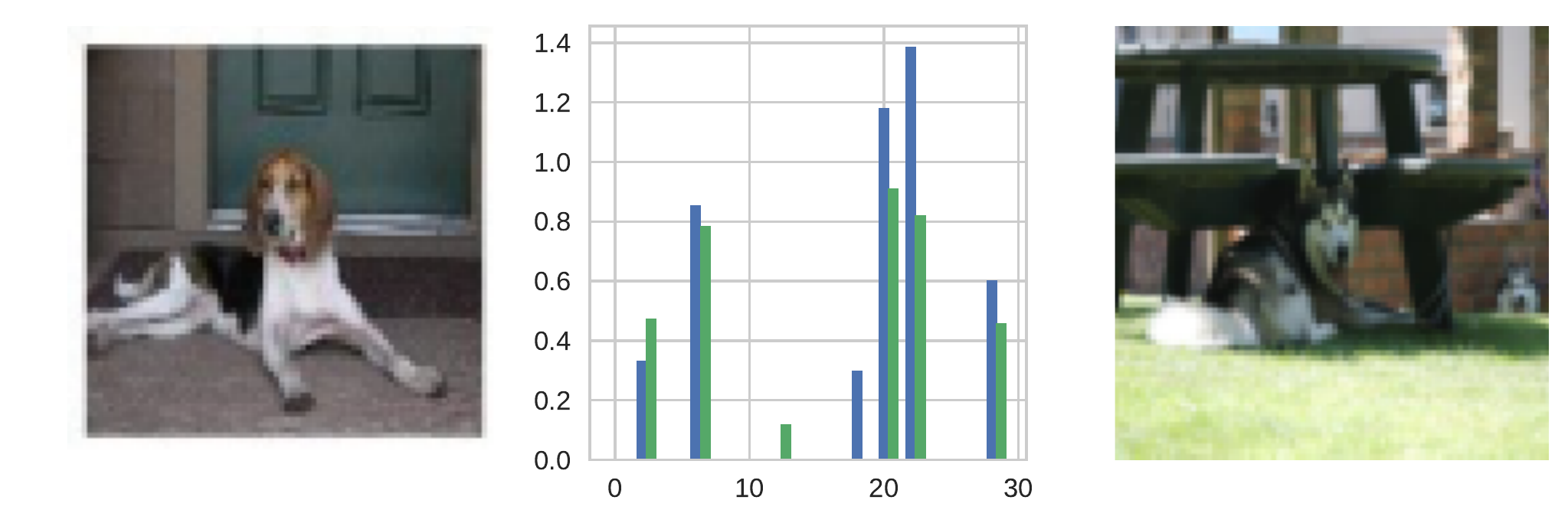}
\includegraphics[width=0.48\textwidth]{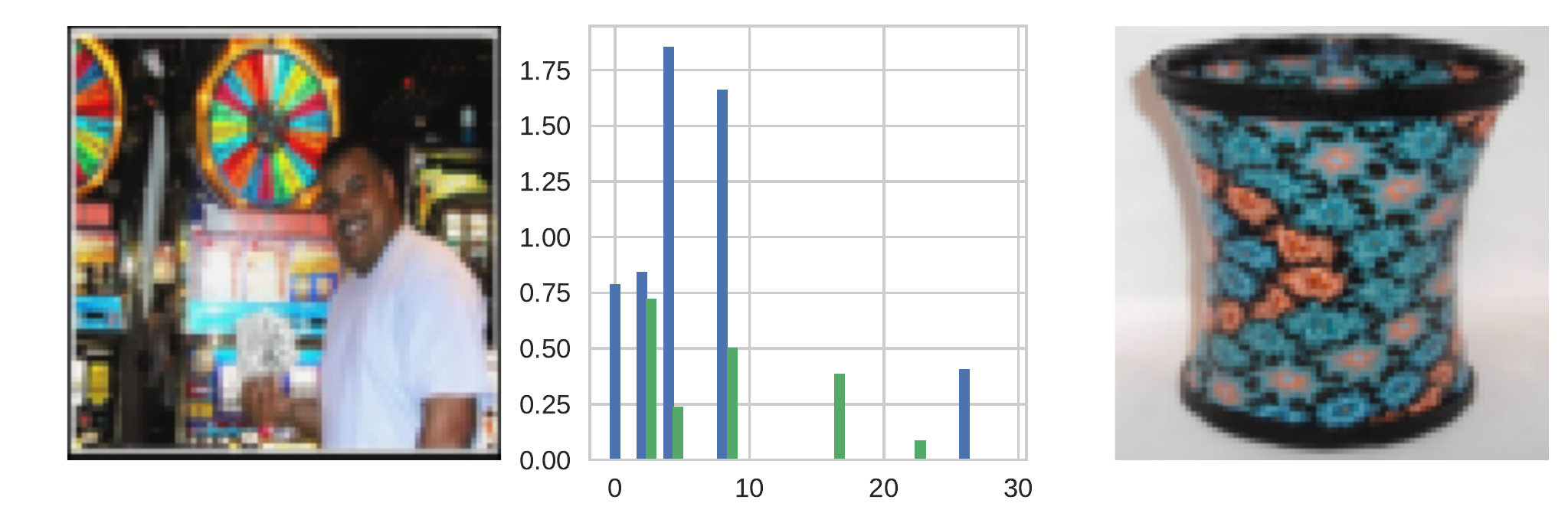}
\caption{\small{After sampling two datasets $D \in\omathcal{D}_{\rm tr}$ and $D'\in\omathcal{D'}_{\rm test}$, we show on the left the two images $x\in D,\; x'\in D$  that minimize $||h(x, \lambda) - h(x', \lambda)||$ and on the right the ones that maximize it. In between each of the two couples we compare a random subset of components of $h(x, \lambda)$ (blue) and $h(x', \lambda)$ (green).} \label{fig:repr}}
\end{figure}

Our meta-model design involves the learning of a cross-episode intermediate representation. We design a meta-representation $h$ as a four layers convolutional neural network, where each layer is composed by a convolution with 32 filters, a batch normalization followed by a ReLU activation and a 2x2 max-pooling. The ground models $g^j$ are logistic regressors that take as input the output of $h$. 
Ground models parameters $w^j$ are initialized to $0$ and optimized by few gradient descent steps 
%(5 in the experiment) 
on the cross-entropy loss computed on $D^j_{\operatorname{tr}}$ (note that, fixing $\lambda$, the inner loss is convex with respect to $w^j$). The step-size $\eta$ is also learned.
For each task the final classification model is thus given by the composition of the meta-learner with the ground learner so that the prediction for an input sample $x$ is equal to $g^j(h(x, \lambda), w_T^j)$. 
We highlight that, unlike in \citep{finn_model-agnostic_2017}, the weight of the representation mapping $\lambda$ are kept constant for each episode, and learned across datasets by minimizing the outer objective $f(\lambda)$ in \eqref{eq:general:l2l:form1}.  
 We compute a stochastic gradient of $f(\lambda)$ by sampling mini-batches of 4 episodes and use Adam with decaying learning rate as optimization method for the meta-model variables $\lambda$. Finally we perform early stopping during meta-training and optimize the number of gradient descent updates (see Figure \ref{fig:updates}) based on the mean accuracy on the test sets of episodes in $\omathcal{D}_{\rm{val}}$. We report results in Table \ref{tab:results}. The proposed method, called \emph{Hyper-Representation}, achieves a competitive result despite its simplicity, highlighting the relative importance of learning a good representation independent from specific tasks, on the top of which simple logistic classifiers can perform and generalize well. Figure \ref{fig:repr} provides a visual 
%validation
example  % COMMENTO better words???
of the goodness of the learned representation, showing that examples --- the first form the training, the second from the testing meta-datasets --- from similar classes (different dog breeds) are mapped near by $h$ and, conversely, samples from dissimilar classes are mapped afar. In Appendix \ref{sec:variants} we empirically show the importance of learning $h$ with the proposed framework.
% \vspace{-.1cm}
\vspace{-.4cm}
%
% \vspace{-.2cm}
%

Ongoing experiments aim at combining the first and the second design patterns outlined in Section 4 
%approach of \citep{finn_model-agnostic_2017} with ours 
both in depth (lower layers weights are hyperparameters and higher layers weights initial points) and in width (a portion of filters constitutes the meta-representation, while the weights relative to the rest of filters are considered initialization), and at experimenting with the third pattern. %Furthermore, we are working on  at parametrizing the update rule (meta-optimizer) and at learning a meta-loss function adapted to the specific setting of few-shots learning. 
Moreover we plan to explore settings in which different datasets come form various domains (e.g. visual, natural language, speech, etc.), are linked to diverse tasks (e.g. classification, localization, segmentation, generation and others) and have structurally different instance spaces. 
% (e.g. object classification, location, .. ?)

\section{Conclusions}
We observed that hyperparameter optimization and learning-to-learn share the same mathematical structure, captured by a bilevel programming problem, which consists in minimizing an outer objective function whose value implicitly depends on the solution of
% , are minimizer of
an inner problem. The objective function of this latter problem --- whose optimization variables are identified with the parameters of (ground) models --- is, in turn, parametrized by the outer problem variables, identified either as hyperparameters or parameters of a meta-model, depending on the context.
%  In this latter problem, the variables are identified with (ground) models parameters and the objective function is parameterized by the outer variables
Since the solution of the inner optimization problem does not have, in general, a closed form expression, 
% we approximate the bilevel program 
we formulate a related constrained optimization problem 
by considering explicitly an optimization dynamics for the inner problem (e.g. gradient descent). 
In this way we are able to (A) compute the outer objective and optimize it by gradient descent and (B) 
optimize also variables that parametrize 
% LUCA OPPURE
% include in the variables of the outer problem parameters that appear inside
the learning dynamics. 
% This formulation 
We discussed examples of the framework and present experiments on few-shots learning, introducing a method for learning a shared, cross-episode, representation.

\bibliographystyle{apalike}
\bibliography{references-mp,Learning.To.Learn,Hyperparameters}

% \newpage

\appendix

\section{On variants of representation learning methods} \label{sec:variants}

We report in Table \ref{tab:app:representation} additional results on a series of experiments for one-shot learning on MiniImagenet with the aim of comparing out method for learning a meta-representation outlined in Sections \ref{sec:l2l} and \ref{sec:exp} with other methods for learning representations that involve the factorization of a classifier as $g^j\circ h$. The representation mapping $h$ is either pretrained on the classification problem with all the images in the training meta-dataset or learned with different meta-learning algorithms. 
% from classical literature \citep{baxter1995learning}.
In all the experiments, for each episode $g^j$ is a multinomial logistic regressor learned with few iterations of gradient descent as described in Section \ref{sec:exp}.
% , while the representation mapping $h$ is obtained either  from transfer learning \cite{} \luca{E' giusto paralre di transfer learning? citazione?} or form classical literature in learning-to-learn \citep{baxter1995learning}. 
%  summarize the results.
% the mean classification accuracy for one-shot learning on MiniImagenet.

\begin{table}[h]
\begin{tabular}{lc|lc}
      \hline
  %     \abovespace\belowspace
      Method & Accuracy 1-shot & Method  &  Accuracy 1-shot \\ \hline \hline 
      \emph{NN-conv}          & $39.97$  & 
      	\emph{Bilevel-train 1x5} 	& 	$27.36$				\\
      \emph{NN-linear}				& $41.50$ & 
      	\emph{Bilevel-train 16x5}  & $29.63$				\\
      \emph{NN-softmax} 			& $41.36$ &  	
      	\emph{Approx-train 1x5}    & $24.74$				\\
      \emph{Multiclass-conv} 		& $36.57$	&
      	\emph{Approx-train 16x5} 	& 		$38.80$			\\
      \emph{Multiclass-linear}			& $43.02$ % $43.02 \pm 0.76$
      & 
      	\emph{Classic-train 1x5}		& $24.70$ 			\\
      \emph{Multiclass-softmax} 			& $37.60$ & 
      \emph{Classic-train 16x5}		&  $40.46$ % $40.46\pm 1.60$
				\\
        \hline
\end{tabular}
\caption{ \small Performance of various methods where the representation is either transfered from models trained in a standard multiclass supervised learning setting (left column) or learned in a meta-learning setting (left column). \label{tab:app:representation}}
\end{table}

% \emph{Approx-validation 1x5, 15x5}  $41.12$
In the experiments in the left column we use as representation mapping $h$ the outputs of different layers of two distinct neural networks (denominated \emph{NN} and \emph{Multiclass} in the Table) trained with a standard multiclass supervised learning approach on the totality of examples contained in the training meta-dataset (600 examples for each of the 64 classes
\footnote{We hold-out 3840 uniformly drawn samples to form a small test set. 
}). 
% For each episode we learn a multinomial logistic regressors on the top of $h$ with 5 iterations of gradient descent as described in Section \ref{sec:exp}. 
The first network \emph{NN}, which has 64 filters per layer, achieves a test accuracy of $43.41\%$. It is the same network used to reproduce the \emph{Nearest-neighbour} baseline in Table \ref{tab:results} 
%(see Table \ref{tab:results} and \citep{ravi_optimization_2016})
and it has been trained with an early stopping procedure on the nearest-neighbour classification accuracy computed on episodes sampled from the validation meta-dataset. The network \emph{Multiclass}, which has 32 filters per layer, has instead been trained with an early stopping procedure on the accuracy on a small held-out validation set. Achieving a test accuracy of $46.33$, this second model is superior on the (standard) multiclass classification problem. For each of the network we report experiment using as representation different layers. Specifically:
  \begin{itemize}
  \item \emph{conv}: we use the output of the last convolutional layer as representation, that is $h(x)\in\mathbb{R}^{2304}_+$ for \emph{NN} and $h(x)\in\mathbb{R}^{1152}_+$ for \emph{Multiclass};
  \item \emph{linear}: we use as representation the linear output layer (before applying the softmax operation), so that $h(x)\in\mathbb{R}^{64}$. 
  \item \emph{softmax}: the representation is given by the probability distribution output of the network; in this case $h(x)\in(0,1)^{64}$
  \end{itemize}
 The \emph{linear} representation yields the best result for both of the networks and in the case of \emph{Multiclass} achieves comparable results with previously proposed meta-learning methods.
 
The experiments in the right column, where $h$ is learned with meta-learning techniques, span in two directions: the first one is that of verifying the impact of various approximations on the computation of the hypergradient, and the second one is to empirically assess the importance of the training/validation splitting of each training episode. 
% \luca{Specificare che $h$ e' la solita CNN?} 
In the experiments denoted \emph{Bilevel-train}, we use a bilevel approach but, unlike in section \ref{sec:exp}, we optimize the parameter vector $\lambda$ of the representation mapping by minimizing the loss on the training sets. The outer objective is thus given by
 $$f(\lambda) = 
    \sum\limits_{j=1}^{N}
    \frac{1}{|D^j_{\operatorname{tr}}|} 
    \sum\limits_{z\in D^j_{\operatorname{tr}}} 
    E^j(w^j_T, \lambda, z).$$
We consider episodes with training set composed by 1 and 16 examples per class, denoted \emph{(1x5)} and \emph{(16x5)} respectively. In these cases $f(\lambda)$ goes quickly to 0 
% \luca{(aggiungiamo un plot?)} 
and the learning ceases after few hundred iterations. In \emph{Approx} experiments we consider an approximation of the hypergradient $\nabla f(\lambda)$ by  disregarding the optimization dynamics of the inner objectives (i.e. we set $\partial_{\lambda} w^j_T = 0$).
% \luca{(scrivaimo questa cosa con piu' precisione?)}
%  and run it computing the errors on the training sets and on the validation sets (\emph{Approx-validation 1x5, 15x5}).
We also run this experiment considering the training/validation splitting obtaining a final test accuracy of $41.12\%$.
 In the experiments denoted as \emph{Classic} we jointly minimize   
  $$f(\lambda, w^1, \dots, w^N) = 
    \sum\limits_{j=1}^{N}
    \frac{1}{|D^j_{\operatorname{tr}}|} 
    \sum\limits_{z\in D^j_{\operatorname{tr}}} 
    E^j(w^j, \lambda, z).$$
 and treat the problem as a standard multitask learning problem as suggested in \citep{baxter1995learning} (with the exception that we evaluate $f$ a mini-batches of 4 episodes, randomly sampled every 4 gradient descent iterations).
 
 This series of experiments suggest that both the training/validation splitting and the full computation of the hypergradient constitute key factors for learning a good meta-representation. 
 Nevertheless, provided that the training sets contain a sufficient number of examples, also the joint optimization method achieves decent results, 
 while learning the representation using only the training sets of one-shot episodes (experiments \emph{train 1x5}) proves unsuccessful in every tested setting, a result
 \footnote{It remains interesting to explore both theoretically and empirically how does the size of validation sets of meta-training episodes impacts on the generalization performances of meta-learning algorithms.}
 in line with the theoretical analysis in \citep{baxter1995learning}. On the other side, using pretrained representations, specially in a low-dimensional space, turns out to be a rather effective baseline. One possible explanation is that, in this context, some classes in the training and testing meta-datasets are rather similar (e.g. various dog breeds) and thus ground classifiers can leverage on very specific representations.

\end{document}